\pdfoutput=1

\documentclass[11pt]{article}

\usepackage{ACL2023}

\usepackage{times}
\usepackage{latexsym}
\usepackage{amsfonts}
\usepackage{amsmath}
\usepackage{booktabs}
\usepackage{graphicx}
\usepackage{multirow}

\usepackage[T1]{fontenc}

\usepackage[utf8]{inputenc}

\usepackage{microtype}

\usepackage{inconsolata}

\title{Sequential Integrated Gradients: a simple but effective method for
explaining language models}

\author{Joseph Enguehard \\
  Babylon Health \\
  Skippr \\
  \texttt{joseph@skippr.com}}

\begin{document}
\maketitle
\begin{abstract}
Several explanation methods such as Integrated Gradients (IG) can be characterised as path-based methods, as they rely on a straight line between the data and an uninformative baseline. However, when applied to language models, these methods produce a path for each word of a sentence simultaneously, which could lead to creating sentences from interpolated words either having no clear meaning, or having a significantly different meaning compared to the original sentence. In order to keep the meaning of these sentences as close as possible to the original one, we propose Sequential Integrated Gradients (SIG), which computes the importance of each word in a sentence by keeping fixed every other words, only creating interpolations between the baseline and the word of interest. Moreover, inspired by the training procedure of several language models, we also propose to replace the baseline token "pad" with the trained token "mask". While being a simple improvement over the original IG method, we show on various models and datasets that SIG proves to be a very effective method for explaining language models.\footnote{An implementation of this work can be found at \url{https://github.com/josephenguehard/time_interpret}}
\end{abstract}

\section{Introduction}

\begin{figure}[ht]
\includegraphics[width=7cm]{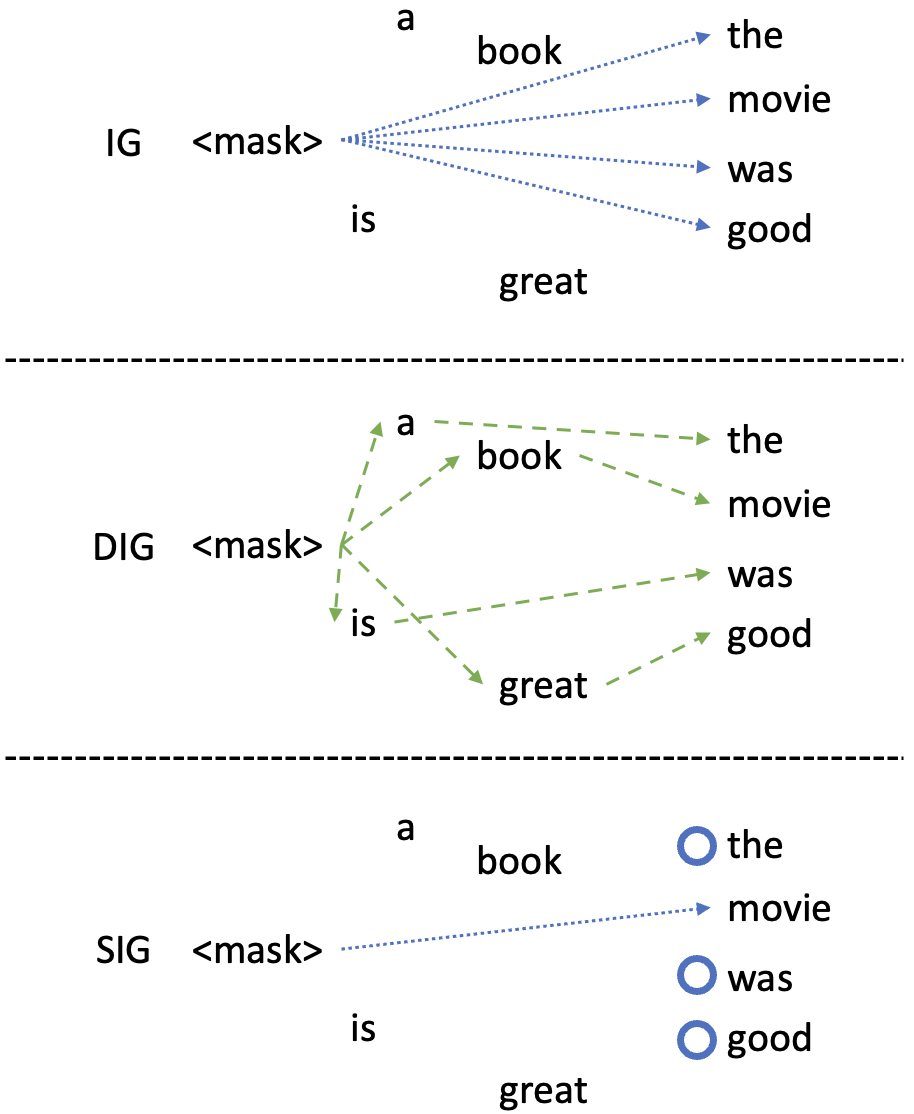}
\centering
\caption{\textbf{Comparison between IG, DIG, and our method: SIG.} While DIG improves on IG by creating discretized paths between the data and the baseline, it can produce sentences with a different meaning compared to the original one. Our method tackles this issue by fixing every word to their true value except one, and moving the remaining word along a straight path (SIG)}
\label{method}
\end{figure}

Language models such as BERT \cite{devlin2018bert} have demonstrated to be effective on various tasks, for instance on sentiment analysis \cite{hoang-etal-2019-aspect}, machine translation \cite{zhu2020incorporating}, text summarization \cite{liu2019fine} or intent classification \cite{chen2019bert}. However, with the increased performance and usage of such models, there has been a parallel drive to develop methods to explain predictions made by these models. Indeed, BERT and its variations are complex models which do not allow a user to easily understand why a certain prediction has been produced. On the other hand, it is important to be able to explain a model's predictions, especially when this model is used to make high-stake decisions, or when there is a risk of a discriminating bias, for instance when detecting hate speech on social media \cite{sap2019risk}.

As a result, developing effective methods to explain not only language models, but also machine learning models in general, has recently gained significant attention. Many different methods have therefore been proposed such as: LIME \cite{ribeiro2016should}, Grad*Inp \cite{shrikumar2016not}, Integrated Gradients (IG) \cite{sundararajan2017axiomatic}, DeepLift \cite{shrikumar2017learning} or GradientShap \cite{lundberg2017unified}. Among these methods, some can be characterised as path-based, which means that they rely on a straight line between the data and an uninformative baseline. For instance, IG computes gradients on interpolated points along such a path, while DeepLift and GradientShap can be seen as approximations of IG \cite{ancona2017towards, lundberg2017unified}.

While these methods aim to be used on any type of models and data, some have been tailored to the specificity of language models. For instance, \citet{sanyal2021discretized} challenge the use of continuous paths on a word embedding space which is inherently discrete. They propose as a result Discretized Integrated Gradient (DIG), which replaces the continuous straight path with a discretized one, where interpolated points are words.

In our work, we suggest another potential issue when applying path-based explanation methods on language models. These models are usually designed to be used on individual or multiple sentences, in order to perform for instance sentiment analysis or question answering. However, a path-based method applied on such models creates straight lines between each word and a baseline simultaneously. When interpolated points are grouped together to form a sentence, this sentence could have a very different meaning compared with the original one.

As a result, we propose a simple method to alleviate this potential issue: computing the importance of each word in a sentence or a text by keeping fixed every other word and only creating interpolations between the baseline and the word of interest. After computing the importance of each word in this way, we normalise these attributions across the sentence or text we aim to explain. We call this method Sequential Integrated Gradients (SIG), as, although we focus in this work on language models, such a method could be used on any sequential modelling. We also propose to use the token "mask" as a baseline, when possible, as its embedding has been trained to replace part of sentences when training language models. As a result, our method follows closely the training procedure of these models.

\section{Method}
\label{sec:method}

\begin{table*}[t]
	\centering
	\resizebox{\textwidth}{!}{%
		\begin{tabular}{lccccccccc}
			\toprule
			\multirow{2}{*}{\textbf{Method}}		& \multicolumn{3}{c}{\textbf{DistilBERT}}	& \multicolumn{3}{c}{\textbf{RoBERTa}} & \multicolumn{3}{c}{\textbf{BERT}}	\\
			\cmidrule(r){2-4} \cmidrule(r){5-7} \cmidrule(r){8-10}
			& LO $\downarrow$ & Comp $\uparrow$ & Suff $\downarrow$ & LO $\downarrow$ & Comp $\uparrow$ & Suff $\downarrow$ & LO $\downarrow$ & Comp $\uparrow$ & Suff $\downarrow$ 	\\
			\midrule
			Grad*Inp				& -0.412 & 0.112	& 0.375	& -0.199	& 0.0760	& 0.426	& -0.263  & 0.0923	& 0.439	\\
			DeepLift				& -0.624 & 0.170	& 0.271	& -0.261	& 0.0932	& 0.408	& -0.244  & 0.0898	& 0.438	\\
			GradientShap			& -1.32	& 0.303	& 0.258	& -0.896	& 0.261	& 0.314	& -0.622 & 0.219  & 0.388  \\
			IG				        & -1.96	& 0.445	& 0.151	& -1.44	& 0.405	& 0.226	& -0.981	& 0.345	& 0.352	\\
			DIG 				    & -1.69	& 0.384	& 0.167	& -0.824	& 0.263	& 0.278	& -0.777	& 0.287	& 0.345	\\
			\midrule
			SIG 	  				& \textbf{-2.02}	& \textbf{0.473}	& \textbf{0.0992}	& \textbf{-1.62}	& \textbf{0.440}	& 	\textbf{0.216} &  \textbf{-1.19}  	& \textbf{0.392}	& \textbf{0.312}	\\
			\bottomrule
		\end{tabular}%
	}
	\caption{Comparison of SIG with several feature attribution methods on three language models fine-tuned on the SST2 dataset. For $\uparrow$ metrics, the higher the better, while for $\downarrow$ ones, the lower the better.}
	\label{tab:results_sst}
\end{table*}

\begin{table*}[t]
	\centering
	\resizebox{\textwidth}{!}{%
		\begin{tabular}{lccccccccc}
			\toprule
			\multirow{2}{*}{\textbf{Method}}		& \multicolumn{3}{c}{\textbf{DistilBERT}}	& \multicolumn{3}{c}{\textbf{RoBERTa}} & \multicolumn{3}{c}{\textbf{BERT}}	\\
			\cmidrule(r){2-4} \cmidrule(r){5-7} \cmidrule(r){8-10}
			& LO $\downarrow$ & Comp $\uparrow$ & Suff $\downarrow$ & LO $\downarrow$ & Comp $\uparrow$ & Suff $\downarrow$ & LO $\downarrow$ & Comp $\uparrow$ & Suff $\downarrow$ 	\\
			\midrule
			Grad*Inp				& -0.153	& 0.0766	& 0.209	& -0.0892	& 0.0432	& 0.300	& -0.291	& 0.0887	& 0.298	\\
			DeepLift				& -0.269	& 0.117	& 0.159	& -0.124	& 0.0557	& 0.269	& -0.285	& 0.0701	& 0.366	\\
			GradientShap		    & -0.832	& 0.289	& 0.137	& -0.606	& 0.204	& 0.144	& -0.874	& 0.172	& 0.308	\\
			IG				        & -1.50	& 0.534	& 0.0428	&  -1.35	& \textbf{0.441}	& 0.0327	& -1.58	& 0.302 & 0.224	\\
			DIG				        & -0.779	& 0.304	& 0.133	& -0.663	& 0.186	& 0.108	& -1.06	& 0.207	& 0.232	\\
			\midrule
			SIG				        & \textbf{-1.95}	& \textbf{0.564}	& \textbf{0.00409}	&  \textbf{-1.37}	& 0.404	& \textbf{-3.31E-05} & \textbf{-2.12}	& \textbf{0.364}	& \textbf{0.124}	\\
			\bottomrule
		\end{tabular}%
	}
	\caption{Comparison of SIG with several feature attribution methods on three language models fine-tuned on the IMDB dataset.}
	\label{tab:results_imdb}
\end{table*}

\paragraph{SIG formulation}

Let's define a language model as a function $\textrm{F}(\textbf{x}): \mathbb{R}^{m \times n} \rightarrow \mathbb{R}$. The input \textbf{x} is here modelled as a sequence of $m$ words, each having $n$ features. These features are usually constructed by an embedding layer. We denote $\textbf{x}_i$ the $\textrm{i}^{\textrm{th}}$ word of a sentence (or of a text, depending on the input of the model), and $x_{ij}$ the $\textrm{j}^{\textrm{th}}$ feature of the $\textrm{i}^{\textrm{th}}$ word. The output of F is a value in $\mathbb{R}$, which is, in our experiments, a measure of the sentiment for a given sentence. We now define the baseline for each word $\textbf{x}_i$ as $\overline{\textbf{x}}^i = (\textbf{x}_1, ..., \texttt{<mask>}, ..., \textbf{x}_m) $. The baseline is therefore identical to $\textbf{x}$ except at the $\textrm{i}^{\textrm{th}}$ position, where the word $\textbf{x}_i$ is replaced by the embedding of the word "mask"\footnote{Certain language models, such as GPT-2 \citep{radford2019language}, do not have a "mask" token. A "pad" token should be therefore used for such models.}, a token used in many language model to replace part of the sentence during training. Moreover, we use the notation $\overline{\textbf{x}}^i$ instead of $\overline{\textbf{x}}_i$ as $\overline{\textbf{x}}^i$ corresponds to an entire sentence, not to be mistaken with a single word like $\textbf{x}_i$.

In this setting, we keep the baseline as similar to the original sentence as possible, only changing the word of interest. This method of explaining a word is also kept similar to the way these language models are usually pre-trained, by randomly masking part of sentences. 

Let's now define our Sequential Integrated Gradients (SIG) method. For a word $\textbf{x}_i$ and a feature $j$, SIG is defined as:
\begin{multline*}
    \textrm{SIG}_{ij}(\textbf{x}) := (x_{ij} - \overline{x}_{ij}) \times \\ \int_0^1 \frac{\partial \textrm{F}(\overline{\textbf{x}}^i + \alpha \times (\textbf{x} - \overline{\textbf{x}}^i))}{\partial x_{ij}} \, d\alpha
\end{multline*}

Similar to the original IG \cite{sundararajan2017axiomatic}, we compute the gradient of F along a straight line between $\overline{\textbf{x}}^i$ and $\textbf{x}$ for each word $\textbf{x}_i$, the main difference being that the baseline differs for each word. Also similar to the original IG, we approximate in practice the integral with Riemann summation.

Finally, we compute the overall attribution of a word by computing the sum over the feature dimension j, and normalising the result:

$$ \textrm{SIG}_i(\textbf{x}) := \frac{\sum_j \textrm{SIG}_{ij}}{||\textrm{SIG}||} $$

\paragraph{Axioms satisfied by SIG}
\label{sec:method_axioms}

The original Integrated Gradients method satisfies a few axioms that are considered desirable for any explanation methods to have. Among these axioms, SIG follows implementation invariance, which states that attributions should be identical if two models are functionally equivalent. Moreover, SIG follows completeness in a specific way: for each word $\textbf{x}_i$, we have the following result:

$$\sum_j SIG_{ij}(\textbf{x}) =  \textrm{F}(\textbf{x}) - \textrm{F}(\overline{\textbf{x}}^i)$$

This means that for each word, the sum of its attribution across all features j is equal to the difference between the output of the model as $\textbf{x}$ and at its corresponding baseline $\overline{\textbf{x}}^i$. However, it does \textbf{not} entail that $\sum_{ij} SIG_{ij}(\textbf{x}) = \textrm{F}(\textbf{x}) - \textrm{F}(\overline{\textbf{x}})$, where $\overline{\textbf{x}}$ would be an overall baseline filled with $\texttt{<mask>}$.

Moreover, this last axiom entail another one called sensitivity, which here means that if, for a certain word, the input $\textbf{x}$ has the same influence on the output of F as its corresponding baseline $\overline{\textbf{x}}^i$, then $\sum_j SIG_{ij}(\textbf{x}) = 0$.

Finally, we show in Appendix \ref{app:symmetry} that SIG preserves symmetry for each word on the embedding dimension, but that this axiom is not true in general.

\paragraph{Using mask instead of pad as a baseline}
\label{sec:mask_pad}

We propose in this study to replace, as the baseline, the commonly used "pad" token with the "mask" token, on language models having such token. This seems to go against the intuition that the baseline should be uninformative, as "mask" is a trained token. To support the usage of "mask", we argue that, because \texttt{<PAD>} (denoting the embedding of "pad") is untrained, it could be arbitrarily close to some words, and far from others. Oh the other hand, \texttt{<MASK>} has been trained to replace random words, making it ideally as close to one word as to any other.

Another way to see it is to compare it with images. It is natural for images to choose the baseline as a black image, as this baseline has no information.
However, there is no such guarantee in NLP. For instance, the embedding of "pad": \texttt{<0, 0, 0, \dots, 0>} could perfectly be very close to an embedding of a word with a specific meaning, which would harm the explanations. On the other hand, \texttt{<MASK>} has been trained to replace any word, and therefore seems more suited to be the baseline.
\section{Experiments}

\begin{table*}[t]
	\centering
	\resizebox{\textwidth}{!}{%
		\begin{tabular}{lccccccccc}
			\toprule
			\multirow{2}{*}{\textbf{Method}}		& \multicolumn{3}{c}{\textbf{DistilBERT}}	& \multicolumn{3}{c}{\textbf{RoBERTa}} & \multicolumn{3}{c}{\textbf{BERT}}	\\
			\cmidrule(r){2-4} \cmidrule(r){5-7} \cmidrule(r){8-10}
			& LO $\downarrow$ & Comp $\uparrow$ & Suff $\downarrow$ & LO $\downarrow$ & Comp $\uparrow$ & Suff $\downarrow$ & LO $\downarrow$ & Comp $\uparrow$ & Suff $\downarrow$ 	\\
			\midrule
			Grad*Inp				& -0.257	& 0.0681	& 0.315	& -0.121	& 0.0617	& 0.363	& -0.438	& 0.143	& 0.438	\\
			DeepLift				& -0.332	& 0.101	& 0.260	& -0.163	& 0.0804	& 0.348	& -0.452	& 0.123	& 0.450	\\
			GradientShap			& -0.452	& 0.237	& 0.212	& -0.389	& 0.194	& 0.299	& -0.715	& 0.204	& 0.438	\\
			IG				        & \textbf{-0.540}	& \textbf{0.341}	& 0.163	& -0.787	& 0.354	& \textbf{0.242}	& -1.19	& 0.307	& 0.410	\\
			DIG      				& -0.487	& 0.273	& 0.181	& -0.426	& 0.223	& 0.286	& -1.05	& 0.293	& 0.414	\\
			\midrule
			SIG      				& -0.533	& 0.331	& \textbf{0.134}	& \textbf{-0.869}	& \textbf{0.361}	& 0.251	& \textbf{-1.52}	& \textbf{0.390}	& \textbf{0.349}	\\
			\bottomrule
		\end{tabular}%
	}
	\caption{Comparison of SIG with several feature attribution methods on three language models fine-tuned on the Rotten Tomatoes dataset.}
	\label{tab:results_rotten}
\end{table*}

\begin{table}[t]
	\centering
	\resizebox{0.48\textwidth}{!}{%
	\begin{tabular}{lcccccc}
		\toprule
		& steps & LO $\downarrow$ & Comp $\uparrow$ & Suff $\downarrow$ & Delta & Time \\
		\midrule
		IG & 50     & -0.981  & 0.345	& 0.352 & 0.304 & $t$	\\
		SIG & 50    & \textbf{-1.19}	& \textbf{0.392}	& \textbf{0.312} & 4.82 & N $\times$ $t$	\\
		\midrule
		IG & 250     & -0.999    & 0.352 & 0.355 & 0.055 & $t'$ \\
        IG & 10 $\times$ N     & -0.998   & 0.351 & 0.352 & 0.066 & N $\times$ $t'$ / 25  \\
		SIG & 10    &  \textbf{-1.14}   & \textbf{0.373} & \textbf{0.322} & 4.93 & N $\times$ $t'$ / 25 \\
		\bottomrule
	\end{tabular}%
	}
	\caption{Comparison of IG and SIG with different numbers of interpolations on BERT fine-tuned on the SST2 dataset. 
    \\
    $t$ and $t'$ represent the amount of time to calculate IG with 50 and 250 steps respectively, and N represents the number of words on the input data (for instance in one sentence). On the SST2 dataset, we have an average of: N $\approx 25$ words per sentence.
    \\
    On top of the table, we compare IG and SIG using a fixed number of steps.
    On the bottom of the table, we compare IG with 250 steps against SIG with 10 steps. Since N $\approx 25$, we have N $\times$ $t'$ / 25 $\approx$ $t'$. For a fairer comparison, we also compare IG with a variable number of steps: 10 $\times$ N for each sentence, against SIG with 10 steps. These two methods have the same time complexity.
    \\
    Delta is defined as $\sum_{ij} Attr_{ij}(\textbf{x}) - (\textrm{F}(\textbf{x}) - \textrm{F}(\overline{\textbf{x}}))$. Contrary to IG, SIG has a high delta value, as in general $\sum_{ij} SIG_{ij}(\textbf{x}) \neq \textrm{F}(\textbf{x}) - \textrm{F}(\overline{\textbf{x}})$.
 }
	\label{tab:results_10_steps}
\end{table}

\subsection{Experiments design}
\label{"sec:exp_design}

We evaluate SIG against various explanation methods by closely following the experimental setup of \citet{sanyal2021discretized}. As such, we use the following language models: BERT \cite{devlin2018bert}, DistilBERT \cite{sanh2019distilbert} and RoBERTa \cite{liu2019roberta}. We also use the following datasets: SST2 \cite{socher2013recursive}, IMDB \cite{maas2011learning} and Rotten Tomatoes \cite{pang2005seeing}, which classify sentences into positive or negative sentiments or reviews. Moreover, we use the HuggingFace library to recover processed data and pretrained models \cite{wolf2019huggingface}.

Following \cite{sanyal2021discretized}, we use the following evaluation metrics: Log-Odds \cite{shrikumar2017learning}, Comprehensiveness \cite{deyoung2019eraser} and Sufficiency \cite{deyoung2019eraser}. These metrics mask the top or bottom 20 \% important features, according to an attribution method, and measure by how much the prediction of the language model changes using this masked data, compared with the original one. For more details on these metrics, please see \citet{sanyal2021discretized}.

Finally, we use the following feature attribution methods to compare our methods against: Grad*Inp \cite{shrikumar2016not}, Integrated Gradients \cite{sundararajan2017axiomatic}, DeepLift \cite{shrikumar2017learning}, GradientShap \cite{lundberg2017unified} and Discretized IG (DIG) \cite{sanyal2021discretized} using the GREEDY heuristics. Moreover, as in \citet{sanyal2021discretized}, we use 50 interpolation steps for all methods expect from DIG, for which we use 30 steps.

\begin{table*}[t]
	\centering
	\resizebox{\textwidth}{!}{%
	\begin{tabular}{lc}
		\toprule
		\textbf{Method} & Example \\
		\midrule
		IG & “a well-made and often \textbf{\underline{lovely}} \textbf{depiction} of the \textbf{mysteries} of friendship. \\
		SIG & “a \textbf{\underline{well-made}} and often \textbf{lovely} depiction of the mysteries of friendship. \\
        \midrule
        IG & “\textbf{a} hideous , \textbf{\underline{confusing}} spectacle , \textbf{one} that may well put \textbf{the} nail in the coffin of any future rice adaptations.” \\
        SIG & “a \textbf{\underline{hideous}} , \textbf{confusing}
        spectacle , \textbf{one} that may well put the nail in the coffin of any future \textbf{rice} adaptations.” \\
        \midrule
        IG & "this is \textbf{\underline{junk}} food cinema at its \textbf{gr}easiest." \\
        SIG & "this is \textbf{\underline{junk}} food cinema at its gr\textbf{ea}siest." \\
        \midrule
        IG & "a \textbf{\underline{remarkable}} 179-minute \textbf{meditation} on the nature of revolution." \\
        SIG & "a \textbf{\underline{remarkable}} 179-minute \textbf{meditation} on the nature of revolution." \\
		\bottomrule
	\end{tabular}%
	}
	\caption{Examples of attributions on several sentences of the SST2 dataset. The \textbf{\underline{underlined bold}} tokens represent the most important token in the sentence, while \textbf{bold} tokens represent the top 20 \% tokens in the sentence, according to each attribution method.}
	\label{tab:examples}
\end{table*}

\subsection{Results}

\paragraph{Comparison with other feature attribution methods}
We present of Tables \ref{tab:results_sst}, \ref{tab:results_imdb} and \ref{tab:results_rotten} a comparison of the performance of SIG with the attribution methods listed in \ref{"sec:exp_design}. We observe that SIG significantly outperforms all other methods across most datasets and language models we used. This tends to confirm that the change of overall meaning of a sentence by combining interpolations simultaneously is an important issue which needs to be tackled.

\paragraph{Comparison between IG and DIG}

Although results in \citet{sanyal2021discretized} show that DIG outperforms other methods, including IG, this is not the case when using "mask" as a token. This result seems to undermine the intuition of \citet{sanyal2021discretized} that the discrete nature of the embedding space is an important factor when explaining a language model. We also show in Appendix \ref{app:non_monotonic} that the requirement of having a monotonic path, stressed by \citet{sanyal2021discretized}, is not necessary.

\paragraph{Choice of the baseline token}

We also provide in Appendix \ref{app:pad} results using "pad" as a baseline. Comparison between Tables \ref{tab:results_sst}, \ref{tab:results_imdb} and \ref{tab:results_rotten} on one hand, and Tables \ref{tab:pad_results_sst}, \ref{tab:pad_results_imdb}, \ref{tab:pad_results_rotten} on the other hand show that IG greatly improves using the "mask" token as a baseline. This seems to confirm our intuition of using this token instead of "pad". Moreover, SIG performs similarly using either token, which demonstrates the robustness of this method across these two baseline tokens.

\paragraph{Time complexity of SIG}
\label{sec:exp_time_comp}

One important drawback of SIG is its time complexity, which is dependent on the number of words in the input data. In Table \ref{tab:results_10_steps}, we compare the original IG with SIG, using different numbers of steps. We define $t$ and $t'$ as the time complexity of computing IG with respectively 50 and 250 steps, and N the number of words in the input data. This table shows that, although reducing the number of steps results in a decrease of performance, SIG with 10 steps still performs better than both IG with 250 steps and IG with 10 $\times$ N steps, while having the same time complexity. 

Moreover, as noted in \citet{sanyal2021discretized}, using IG with a large number of steps decreases Delta = $\sum_{ij} IG_{ij}(\textbf{x}) - (\textrm{F}(\textbf{x}) - \textrm{F}(\overline{\textbf{x}}))$, while not improving performance. As a result, when computing attributions on long sentences or large texts, we recommend using SIG with a reduced number of steps instead of IG. 

\paragraph{Comparison of IG and SIG on several examples}
\label{sec:examples}

We provide on Table~\ref{tab:examples} several examples of explained sentences, using IG and SIG. Both methods tend to agree on short sentences, while more disagreements appear on larger ones. For each example, we display in underlined bold the most important token, and in bold the top 20 \% most important tokens, according to each method.

\section{Conclusion}

In this work, we have defined an attribution method specific to text data: Sequential Integrated Gradients (SIG). We have shown that SIG yields significantly better results than the original Integrated Gradients (IG), as well as other methods specific to language models, such as Discretized Integrated Gradients (DIG). This suggests that keeping the meaning of interpolated sentences close to the original one is key to producing good explanations. We have also shown that, although SIG can be computationally intensive, reducing the number of interpolations still yields better results than IG with a greater number of interpolations.

We have also highlighted in this work the benefit of using the token "mask" as a baseline, instead of "pad". Although SIG seems to be robust across both tokens, this is especially important when using IG, as it significantly improves the quality of explanations. Using the trainable token "mask" is indeed closer to the training procedure of language models, and should yield better interpolations as a result. We recommend therefore using this token as a baseline, when possible, when explaining predictions made by a language model.

Moreover, while this study was conducted on bidirectional language models such as BERT, SIG could also be used on auto-regressive models such as GPT-2 \citep{radford2019language}, by iteratively computing the attribution of a token, while keeping previous tokens fixed, and masking future tokens if any has been already computed.
\section*{Limitations}

We see two main limitations of this work. The first one concerns the diversity of the language models and datasets used. BERT, DistilBERT and RoBERTa have similar architecture, and SST2, IMDB and Rotten Tomatoes are datasets designed to evaluate the sentiment of English text. It would therefore be interesting to validate the robustness of our results on more diverse languages, tasks and language models. In this short paper, we decided for brevity to follow the experiment design of \citet{sanyal2021discretized}, while being aware of its inherent limitations.

The second limitation of this work concerns the time complexity of SIG. As it needs to compute explanations for each word individually, this method can become very computationally expensive when applied on large text data. To alleviate this issue, we first made it possible to compute gradients in parallel, using an internal batch size similar to how Captum \cite{kokhlikyan2020captum} implemented the Integrated Gradients method. Secondly, as discussed in \ref{sec:exp_time_comp}, it is possible to reduce the number of interpolated points, which makes the computation faster while retaining better performance than the original IG.

In this work, we ran our experiments on a machine with 16 CPUs, and one Nvidia Tesla T4 GPU. With this setting, computing SIG on SST2 and Rotten Tomatoes takes around one hour for each model. On the larger IMDB, computing SIG, on 2000 randomly sampled inputs, takes around 5 days for BERT and RoBERTa, and 2 days for DistilBERT.
\section*{Ethics Statement}

The methods presented in this work aim to explain language models, and can as such present ethical issues related to this task. Discriminating biases can indeed be present in text data on which a language model is trained, and such a model can acquire and propagate these biases \cite{sap2019risk}. As the presented methods aim to explain a language model without additional knowledge, these methods could also display discriminating biases learnt by a language model.

Moreover, common explanation methods such as Integrated Gradients has proved to be prone to adversarial attacks \cite{dombrowski2019explanations}, and can be misleading when used on out of sample data \cite{slack2021counterfactual}. There is no reason to believe our methods would be more robust compared to existing methods such as IG.

The proposed methods can also be characterised as gradient-based, as they rely on computing gradients on the input data, an uninformative baseline, or on interpolated points between them. As noted by \cite{mittelstadt2019explaining}, such methods are only local and may not give a clear explanation of the model globally.
\section*{Acknowledgement}

The author would like to thank Vitalii Zhelezniak for his thoughtful comments and suggestions, including using the "mask" token as a baseline. We also thank Anthony Hu for his detailed initial review of this paper.

\bibliographystyle{acl_natbib}
\bibliography{acl2023}

\appendix
\section{On the symmetry-preserving axiom of Sequential Integrated Gradients}
\label{app:symmetry}

This section is divided into two parts. First, we show that SIG preserves symmetry for each word along the embedding dimension. Second, we provide a counterexample to show that symmetry does not hold in general.

\paragraph{Symmetry of $\textrm{SIG}_i$}

Let us use the same notations as in Section \ref{sec:method}. We want to compute the attribution of a word $\textbf{x}_i$ on a model F, using the baseline $\overline{\textbf{x}}^i$. Let's define the function:

$$ \textrm{F}_i (\textbf{x}) := \textrm{F}(\textbf{x}_1, ..., \textbf{x}, ..., \textbf{x}_m) $$

$\textrm{F}_i$ corresponds to F where only the $\textrm{i}^{\textrm{th}}$ word is not fixed. Here, $\textbf{x}$ corresponds to a word, not a sentence. 

For such a function, SIG corresponds to the regular IG method: the baseline is $<\textrm{mask}>$ and SIG constructs a straight line between this baseline and $\textbf{x}_i$. As a result, if $\textrm{F}_i$ is symmetric on two embedding features $j_1$ and $j_2$, SIG preserves this symmetry: $\textrm{SIG}_{ij_1}(\textbf{x}) = \textrm{SIG}_{ij_2}(\textbf{x})$.

\paragraph{Non symmetry of $\textrm{SIG}$}

The fact that SIG does not preserve symmetry in general is due to the choice of the baseline. As a counterexample, let's define a language F which takes as an input two words $\textbf{x}_1$ and $\textbf{x}_2$. This language model is moreover symmetric: $\textrm{F}(\textbf{x}_1, \textbf{x}_2) = \textrm{F}(\textbf{x}_2, \textbf{x}_1)$. 

Here, the original IG method would preserve the symmetry: as the baseline is $(\texttt{<mask>}, \texttt{<mask>})$, when $\textbf{x}_1 = \textbf{x}_2$, we have $IG(\textbf{x})_1 = IG(\textbf{x})_2$. However, SIG doesn't preserve the symmetry due to its baseline: we would have: $\overline{\textbf{x}}^1 = (\texttt{<mask>}, \textbf{x}_2)$ and $\overline{\textbf{x}}^2 = (\textbf{x}_1, \texttt{<mask>})$. As a result, $\textrm{SIG}(\textbf{x})_1 = \textrm{SIG}(\textbf{x})_2$ only if $\textbf{x}_1 = \textbf{x}_2 = \texttt{<mask>}$.

\section{Additional results using the "pad" token}
\label{app:pad}

We present in this section results using the "pad" token instead of the "mask" one. These results for the three datasets: SST2, IMDB and Rotten Tomatoes can be found respectively on Tables \ref{tab:pad_results_sst}, \ref{tab:pad_results_imdb} and \ref{tab:pad_results_rotten}.

When using the "pad" token as a baseline, SIG seems to perform similarly compared with using the "mask" one, while other methods perform significantly worse. This demonstrates both the need to use "mask" as a token, and the robustness of the SIG method across different baselines.

\begin{table*}[t]
	\centering
	\resizebox{\textwidth}{!}{%
		\begin{tabular}{lccccccccc}
			\toprule
			\multirow{2}{*}{\textbf{Method}}		& \multicolumn{3}{c}{\textbf{DistilBERT}}	& \multicolumn{3}{c}{\textbf{RoBERTa}} & \multicolumn{3}{c}{\textbf{BERT}}	\\
			\cmidrule(r){2-4} \cmidrule(r){5-7} \cmidrule(r){8-10}
			& LO $\downarrow$ & Comp $\uparrow$ & Suff $\downarrow$ & LO $\downarrow$ & Comp $\uparrow$ & Suff $\downarrow$ & LO $\downarrow$ & Comp $\uparrow$ & Suff $\downarrow$ 	\\
			\midrule
			Grad*Inp				& -0.402	& 0.112	& 0.375	& -0.318	& 0.085	& 0.398	& -0.454	& 0.092	& 0.439	\\
			DeepLift				& -0.196	& 0.053	& 0.489	& -0.270	& 0.0784	& 0.439	& -0.283	& 0.061	& 0.463	\\
			GradientShap			& -0.753	& 0.191	& 0.328	& -0.514	& 0.146	& 0.386	& -0.471	& 0.146	& 0.425	\\
			IG 						& -0.954	& 0.251	& 0.273	& -0.726	& 0.227	& 0.315	& -0.658	& 0.235	& 0.398	\\
			DIG                  	& -1.222  & 0.310	& 0.237	& -0.812	& 0.249	& 0.287 & -0.879	& 0.292	& 0.374	\\
			\midrule
			SIG 	                & \textbf{-1.993}  & \textbf{0.466}	& \textbf{0.108}	& \textbf{-1.346}	& \textbf{0.398}	& \textbf{0.244}	& \textbf{-1.30}    & \textbf{0.393}	& \textbf{0.331}	\\
			\bottomrule
		\end{tabular}%
	}
	\caption{Comparison of SIG with several baselines on three language models fine-tuned on the SST2 dataset. For $\uparrow$ metrics, the higher the better, while for $\downarrow$ ones, the lower the better.}
	\label{tab:pad_results_sst}
\end{table*}

\begin{table*}[t]
	\centering
	\resizebox{\textwidth}{!}{%
		\begin{tabular}{lccccccccc}
			\toprule
			\multirow{2}{*}{\textbf{Method}}		& \multicolumn{3}{c}{\textbf{DistilBERT}}	& \multicolumn{3}{c}{\textbf{RoBERTa}} & \multicolumn{3}{c}{\textbf{BERT}}	\\
			\cmidrule(r){2-4} \cmidrule(r){5-7} \cmidrule(r){8-10}
			& LO $\downarrow$ & Comp $\uparrow$ & Suff $\downarrow$ & LO $\downarrow$ & Comp $\uparrow$ & Suff $\downarrow$ & LO $\downarrow$ & Comp $\uparrow$ & Suff $\downarrow$ 	\\
			\midrule
			Grad*Inp				& -0.189	& 0.082	& 0.209	& -0.216	& 0.047	& 0.315	& -0.654	& 0.087	& 0.299	\\
			DeepLift				& -0.032	& -0.005    & 0.515	& -0.149	& 0.031	& 0.374	& -0.519	& 0.027	& 0.465	\\
			GradientShap			& -0.315	& 0.117	& 0.302	& -0.351	& 0.110	& 0.213	& -0.622	& 0.088	& 0.358	\\
			IG 						& -0.474	& 0.186	& 0.201	& -0.499	& 0.169	& 0.114	& -0.577	& 0.117	& 0.288	\\
			DIG                  	& -0.812  & 0.297	& 0.153	& -0.626	& 0.187	& 0.099  & -0.971	& 0.192	& 0.229	\\
			\midrule
			SIG 	                & \textbf{-2.157}  & \textbf{0.585}	& \textbf{0.0062}  & \textbf{-0.856}	& \textbf{0.291}	& \textbf{0.0207}	& \textbf{-1.96}	& \textbf{0.352}	& \textbf{0.152}	\\
			\bottomrule
		\end{tabular}%
	}
	\caption{Comparison of SIG with several baselines on three language models fine-tuned on the IMDB dataset.}
	\label{tab:pad_results_imdb}
\end{table*}

\begin{table*}[t]
	\centering
	\resizebox{\textwidth}{!}{%
		\begin{tabular}{lccccccccc}
			\toprule
			\multirow{2}{*}{\textbf{Method}}		& \multicolumn{3}{c}{\textbf{DistilBERT}}	& \multicolumn{3}{c}{\textbf{RoBERTa}} & \multicolumn{3}{c}{\textbf{BERT}}	\\
			\cmidrule(r){2-4} \cmidrule(r){5-7} \cmidrule(r){8-10}
			& LO $\downarrow$ & Comp $\uparrow$ & Suff $\downarrow$ & LO $\downarrow$ & Comp $\uparrow$ & Suff $\downarrow$ & LO $\downarrow$ & Comp $\uparrow$ & Suff $\downarrow$ 	\\
			\midrule
			Grad*Inp				& -0.152	& 0.068	& 0.315	& -0.211	& 0.062 & 0.363	& -0.806	& 0.143	& 0.438	\\
			DeepLift				& -0.077	& 0.017	& 0.372	& -0.198	& 0.056	& 0.370	& -0.457  & 0.076	& 0.474	\\
			GradientShap			& -0.326	& 0.147	& 0.250	& -0.264	& 0.103	& 0.348	& -0.697    & 0.161	& 0.429	\\
			IG 						& -0.424	& 0.208	& 0.190	& -0.360	& 0.151	& 0.312	& -0.795	& 0.201	& 0.414	\\
			DIG                  	& -0.501  & 0.257	& 0.184	& -0.346	& 0.153	& 0.310  & -1.06	& 0.267	& 0.416	\\
			\midrule
			SIG 	                & \textbf{-0.753}  & \textbf{0.378}	& \textbf{0.109}	& \textbf{-0.771}	& \textbf{0.318}  & \textbf{0.266}	& \textbf{-1.55}	& \textbf{0.360}	& \textbf{0.393}	\\
			\bottomrule
		\end{tabular}%
	}
	\caption{Comparison of SIG with several baselines on three language models fine-tuned on the Rotten Tomatoes dataset.}
	\label{tab:pad_results_rotten}
\end{table*}

\section{Challenge of the monotonic assumption of the path}
\label{app:non_monotonic}

\citet{sanyal2021discretized} stipulate that the path between a baseline and an input needs to be monotonic to allow approximating the integral in IG using Riemann summation. However, while this is true for a Riemann integral, it is also possible to approximate the Riemann–Stieltjes integral, which is a generalisation of Riemann integral, and does not need a monotonic path. We define the Riemann–Stieltjes integral of $ f: [\textbf{a}, \textbf{b}] \to \mathbb{R} $ as:

$$ \int_{\textbf{x} = \textbf{a}}^{\textbf{b}} f(\textbf{x}) \, dg(\textbf{x}) $$

where $g: [0, 1] \to [\textbf{a}, \textbf{b}]$ designates a path. Let us define a partition over $[0, 1]$ as $t_k$ such as $0 \leq t_1 \leq ... \leq t_n \leq 1$. We can then approximate the integral with the sum:

$$ \sum_{i=0}^{n - 1} f(g(c_i)) \times [g(t_{i + 1}) - g(t_i)] $$

where $c_i \in [t_i, t_{i + 1}]$. As such, while the partition $t_i, \, i \in \{1, ..., n\}$ needs to be monotonic, the function g does not need to have this constraint. As a result, we could define a path-based IG method as:

$$ IG_{\gamma}(\textbf{x})_i := \int_{\gamma} \frac{\partial \textrm{F}(\textbf{x})}{\partial x_i} \, dx_i $$

where $\gamma$ is not necessarily monotonic. 

\cite{lundstrom2022rigorous} provide more insights on this topic, and in particular show that the implementation invariance, completeness and sensitivity axioms hold for non-monotonic paths.

For this reason, we decided not to include a combination of DIG and SIG in this study. However, an implementation of this method and the corresponding results can be found in the repository published with this paper.

\end{document}